\documentclass[10pt,twocolumn]{article}

\usepackage{isqcmc2023}
\usepackage{graphicx,url}

\usepackage[utf8]{inputenc}

\title{Investigating the usefulness of Quantum Blur}

\author{James Wootton\inst{1} \and Marcel Pfaffhauser\inst{1}}

\address{IBM Quantum -- IBM Research Zurich\\
         S\"aumerstrasse 4 - 8803 - R\"uschlikon - Switzerland
         \email{jwo@zurich.ibm.com, faf@zurich.ibm.com}
}

\begin{document}

\maketitle

\begin{abstract}
Though some years remain before quantum computation can fully outperform conventional computation, it already provides resources that can be used for exploratory purposes in various fields. This includes certain tasks for procedural generation in computer games, music and art. The so-called `Quantum Blur' method represents the first step on this journey, providing a simple proof-of-principle example of how quantum software can be useful in these areas today. Here we analyse the `Quantum Blur' method and compare it to conventional blur effects. This investigation was guided by discussions with the most prominent user of the method, to determine which features were found most useful. In particular we determine how these features depend on the quantum phenomena of superposition and entanglement.
\end{abstract}

\maketitle

\section{Introduction and Motivation}

Procedural generation is a broad field that consists of methods to algorithmically generate any form of content~\cite{pcg}. This content could consist of music, text, images, maps or more. The algorithms used can vary greatly, depending on the type of content being generated and the requirements it must satisfy for a give use case~\cite{pcg}. They can range from the relatively simple -- such as Perlin noise used in map or texture generation~\cite{lagae:10} -- to computationally hard tasks such as constraint satisfaction~\cite{smith:11}.

`Quantum Blur' is a method for manipulating certain types of data using quantum software~\cite{wootton-fdg,capdeville-21}, devised as first step towards showing that quantum computers could be used for procedural generation~\cite{wootton-fdg,wootton-cog,hamido-21}. Specifically, it is applied to data that can be represented as sets of floating point numbers associated with coordinates. It can therefore be applied in a range of cases pertinent for procedural generation, such as height maps and images where the numbers represent the height and brightness for different coordinates, as well as musical scores for which the numbers represent the loudness of a note whose coordinates are its pitch and the beat at which it is played.

The application of Quantum Blur in these areas shows that even current quantum computing resources are sufficient to contribute at the easy end of procedural generation, and starts the journey towards investigating ever more complex uses for quantum computation.

Henceforth, we will primarily describe Quantum Blur in terms of its application to monochrome images. The method begins by converting images into the form of quantum circuits. These are the simplest element of software for quantum computers, analogous to a set of Boolean operations in conventional digital software. The conversion of images into circuits is done via a so-called amplitude encoding, sometimes used in quantum machine learning~\cite{schuld-20}. Manipulation of the image is then achieved by making changes to the circuit, typically by adding further operations. The encoding is defined such that small modifications to the circuit result in small, blur-like changes to the image.

Though blur effects are most similar to what is created by this method, it is not intended as an improvement upon existing methods. Instead it is a unique effect, intended to demonstrate that designing techniques for procedural generation in terms of quantum software can lead to new directions and perspectives.

Initial use of the method was for procedural generation in games~\cite{piispanen:23}, motivated by the way that blur effects can be used in simple methods for procedural terrain generation~\cite{pcg}. All such use cases were by the original creators of the method, and were primarily based on map generation~\cite{wootton-fdg}.

More interesting than these test cases are the uses by third parties, particularly those for commercial purposes. These provide an independent demonstration that the method has been found useful. It has been used for procedural generation within the upcoming game `C.L.A.Y — The Last Redemption' by MiTale, with the idea that the non-intuitive quantum-based generation will add to the sci-fi atmosphere of the game~\cite{clay}. It has also been used as a tool for artists. In particular,  by the artists Roman Lipski~\cite{lipski:21,lipski:23} and Libby Heaney~\cite{libby:19,libby:20} as a way of creating novel artworks through manipulation of existing images.

Lipski held exhibitions of his work based on Quantum Blur in November of 2021~\cite{lipski:21} and February of 2023~\cite{lipski:23}. We spoke with him regarding this work to determine what aspects of the method he found useful. In particular we wished to determine what he felt was unique about the method, especially in comparison to standard blur effects.

Lipski said that he values the fact that there was an obvious connection between the initial and final images. Indeed, the exhibition was based around original paintings by Lipski, with each presented alongside several derived works made by Quantum Blur. Lipski said that he values the fact that the modified images had as much or even apparently more detail than the original. This includes both small-scale detail, as well as a preservation of asymmetries on a larger scale.

These properties stand in contrast to conventional blurs. Though these do create an obvious connection between the initial and final images, they do so the expense of detail and asymmetry.

In this paper we will investigate the unique behaviour of Quantum Blur, motivated in particular by the observations made in discussion with Lipski. We will also investigate the roles played by the uniquely quantum phenomena of superposition and entanglement in creating this behaviour. We also consider its application to musical scores.

\section{Summary of Quantum Blur}

We will briefly summarize the aspects of Quantum Blur that are relevant for this work. When considering the definition of the effect, we will restrict to the case of monochromatic images (or height maps) defined by a set $h$ of values $0 \leq h(x,y) \leq 1$, where the $(x,y)$ are the pixel coordinates.

To encode this image in a quantum state, each pixel coordinate $(x,y)$ is assigned a bit string $s(x,y)$. These are chosen such that neighbouring coordinates correspond to strings that differ on only one bit (i.e., a Hamming distance of 1)~\cite{wootton-fdg}.

Note that this would not be achieved by simply converting the coordinate values to binary, since the lexicographic order this results in does not have the required properties. Instead, a form of reflected binary code~\cite{lucal-59} is used.

The length of the bit strings used is the minimum required to assign a unique string to each coordinate. Specifically, for an image of size $W \times H$, the number of bits (and therefore qubits) required is $n = \left\lceil \log_2 W \right\rceil + \left\lceil \log_2 H \right\rceil$.

Images are then encoded by preparing the following superposition state for the $n$ qubits,
\begin{equation} \label{eq:state}
\left | h \right \rangle = \frac{ \sum_{(x,y)} \sqrt{h(x,y)} \left | s(x,y) \right \rangle} {\sqrt{\sum_{(x,y)} h(x,y)}} .
\end{equation}
Note that the image could equally be encoded with alternate versions of this superposition with arbitrary phases on each term in the superposition.

The decoding of images is done by determining all the probabilities $p(s(x,y))$. This can be done either through emulation of the quantum process, or running it on quantum hardware (repeating many times to sample the probabilities). The resulting probabilities are then taken to be the $h(x,y)$, after rescaling so that the maximum is equal to 1,
\begin{equation} \label{eq:encoding}
h(x,y) = \frac{p(s(x,y))}{\max_{s'} p(s')}.
\end{equation}

Note that the same output image could equally emerge from a quantum superposition or the classical probability distribution with the same probabilities for each bit string. The distinction between these therefore does not come from the final readout, but from the manipulation of the image while it is encoded as a quantum state.

Manipulation of images is performed by adding quantum gates after the preparation step and before the final readout of probabilities. The simplest effect is the one that the method is named after: the blur.

To induce a blur effect, each $h(x,y)$ should be updated to introduce a dependence on the neighbouring points. Given the encoding of points as bit strings, this can be done with a small angle rotation such as \texttt{rx}. To see this, consider a point $(x,y)$ and the points $(x_j,y_j)$ for which the bit strings $s(x,y)$ and $s(x_j,y_j)$ differ only on the single bit $j$. Applying \texttt{rx} to the corresponding qubit has the following effect on the amplitudes,
$$
\sqrt{h(x,y)} \rightarrow i \cos \frac{\theta}{2} \sqrt{h(x,y)} + \sin \frac{\theta}{2} \sqrt{h(x_j,y_j)}.
$$
This translates into an interpolation between the two $h$ values, parameterized by the angle $\theta$.

When \texttt{rx} is applied to all qubits, we can make the following approximation for small $\theta$,
$$
\sqrt{h(x,y)} \rightarrow i \sqrt{h(x,y)} + \frac{\theta}{2} \sum_j \sqrt{h(x_j,y_j)}.
$$
Though this will be precise only for extremely small values of $\theta$, it nevertheless shows that there will be a blurring of the value $h(x,y)$ by performing an interpolation of the corresponding amplitude with the average of the square roots of the values for all the points $(x_j,y_j)$. Given the encoding of coordinates to bit strings, this will include all neighbouring points, as required for a standard blur.

It is similarly possible to show that \texttt{ry} rotations can induce a blur effect for small angles. However, the \texttt{rx} and \texttt{ry} blurs quickly deviate as $\theta$ is increased when images are encoded as in Eq. \ref{eq:state}. The simplest and most potent example is that of a plain white image, for which $h(x,y)=1$ for all points. Given the encoding of Eq. \ref{eq:encoding}, this corresponds to the state $\left| + \right\rangle ^{\otimes n}$, which is the eigenstate of the \texttt{rx} rotation. The image will therefore not be effected by the \texttt{rx} blur, as would be expected for a standard blur effect.

For \texttt{ry} rotations, however, the $\left| + \right\rangle$ states would be rotated towards the $\left| 1 \right\rangle$, resulting in $\left| 1 \right\rangle ^{\otimes n}$ for $\theta=\pi/2$. The resulting image would have a single bright pixel for the coordinate whose string is all \texttt{1}s, and $h(x,y)=0$ otherwise. Even for small angles, the fact that this single pixel result becomes much more probable than all others causes it to dominate the image. This is a clear demonstration of an interference effect, rather than standard blur-like behaviour. However, in this case it is one that leads to unimpressive visuals.

More general images will also include a significant overlap with $\left| + \right\rangle ^{\otimes n}$, and this component will be similarly affected. Therefore, though both the \texttt{rx} and \texttt{ry} approaches depart from the simple blur effect for large $\theta$, this typically occurs much faster and less usefully for the \texttt{ry} blur. The \texttt{rx} blur is therefore typically used when using the standard encoding of states according to Eq. \ref{eq:state}.

Before continuing, let us first consider the reasons for the stark differences seen between the \texttt{rx} rotations and \texttt{ry} rotations. For this, consider the simple case of a uniform image. This corresponds to the quantum state with $|+\rangle$ on all qubits. These are eigenstates of \texttt{rx}, and so such rotations will have no effect. For \texttt{ry} however, a $\theta=\pm\pi/2$ rotation will result in a definite $|0\rangle$ or $|1\rangle$ on all qubits, and so all the brightness will collect into a single pixel. More generally, a large $|+\rangle^{\otimes n}$ component can be expected to be present due to the baseline brightness. This again remains invariant for \texttt{rx} rotations, but \texttt{ry} rotations again cause it to collapse into a single pixel. Even for small angles, the effect means that a single bit string becomes much more probable than most others. This makes the \texttt{rx} rotations much easier to use effectively. Indeed, these are the rotations most commonly employed in current use cases of Quantum Blur.

\section{Quantifying Blur}

To compare effects from Quantum Blur to conventional blur effects, we will need a way to quantify their results. Blur effects are typically known for smoothing and removing detail. As a consequence of this, they also remove features within an image that cause it to be asymmetric. As such we will use measures of asymmetry and detail in order to quantify their effects.

As a measure of difference between two images, we will use the root mean square difference of their pixel values,
$$
d(h,h') = \sqrt{\frac{\sum_{(x,y)} (h(x,y) - h'(x,y))^2}{WH }}.
$$
This will form the basis of our measure of asymmetry.

Within the context of Quantum Blur, there are a natural set of symmetries to analyze. These are the reflections that result from applying a single qubit bit flip gate (\texttt{x} or \texttt{y}) to any qubit. For a measure of asymmetry, we will take the mean difference between each of these flipped images and the original. Specifically,
$$
a(h) = \frac{\sum_j d(h,h_j)}{n},
$$
Here $h_j$ is the image that results from flipping qubit $j$, and $n$ is the number of qubits that represents the image.

To quantify detail, we will consider the difference between each pixel value and its neighbours. For each pixel we find the maximum such difference. The root mean square of all these maximum differences is then used as our measure of detail.

This measure of detail is defined in terms of the pixel coordinates $(x,y)$. From the perspective of the bit strings $s(x,y)$, the maximization over all neighbouring coordinates corresponds to maximizing over a subset of the neighbouring strings. This motivates us to define a corresponding detail for the state, using the differences between values for all neighbouring strings. This is then the root mean square of all the differences,

$$
\max_j (h(x,y) - h(x_j,y_j))^2,
$$

where $(x_j,y_j)$ are coordinates such that the strings $s(x,y)$ and $s(x_j,y_j)$ differ only on bit $j$.

\begin{figure}[htbp]
\begin{center}
\includegraphics[width=0.9\columnwidth]{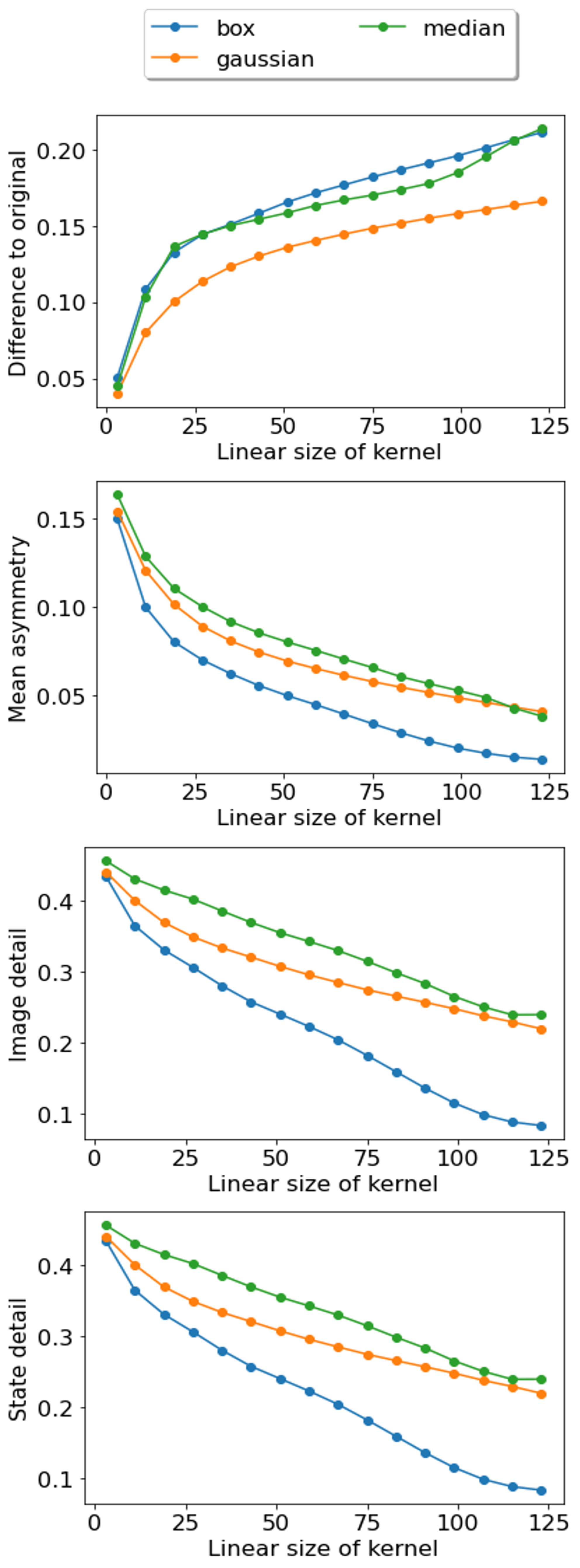}
\caption{Difference to original, asymmetry and detail for various conventional blurs when applied to a $128\times128$ monochrome version of the `sailboat on lake'~\cite{images} test image. The blur effects used here are the box blur, gaussian filtering and median filtering implemented by OpenCV~\cite{opencv_library}}
\label{classical}
\end{center}
\end{figure}

Using these quantities, Fig. \ref{classical} shows the behaviour of classical blur effects, where the strength of the blur is increased by increasing the kernel size. In these results we see that both the asymmetry and detail decrease significantly. As one would expect, the difference between the blurred image and original image also increases. The two measures of detail are found to behave very similarly under such blurs.

\section{The Role of Superposition}

\begin{figure}[htbp]
\begin{center}
\includegraphics[width=0.9\columnwidth]{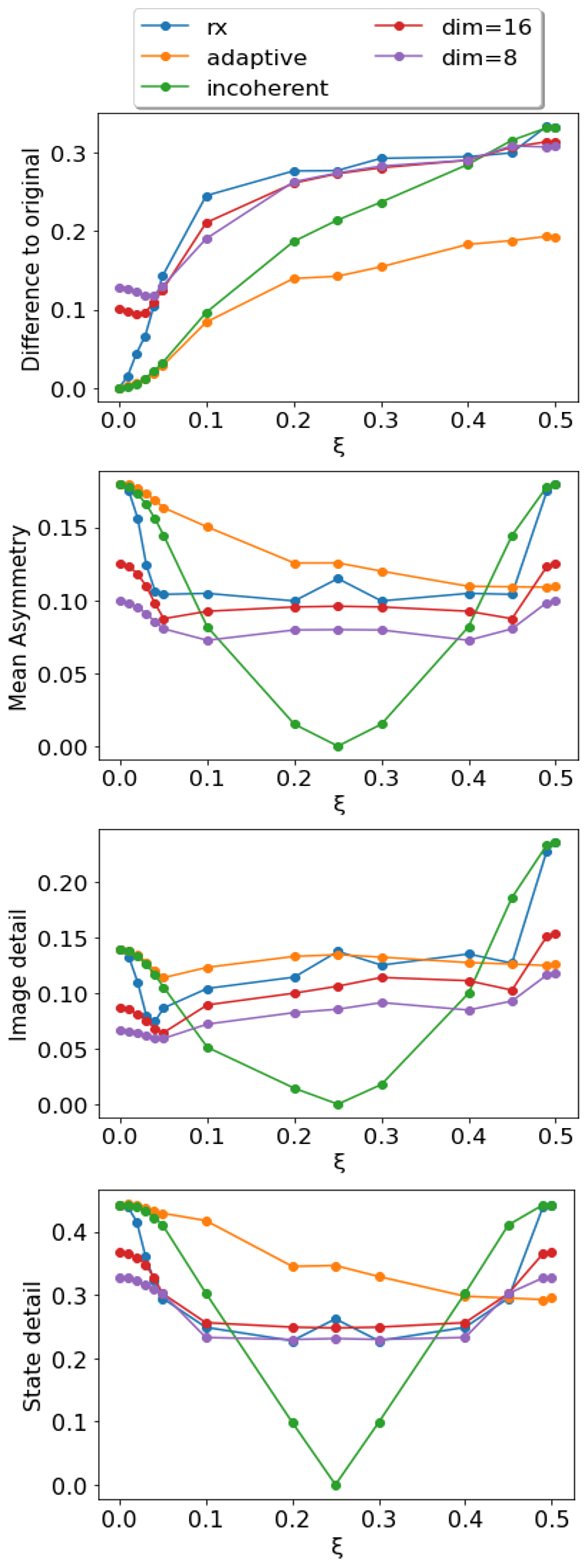}
\caption{Difference to original, asymmetry and detail for a Quantum Blur implemented with \texttt{rx} rotations on each qubit and an incoherent analogue of Quantum Blur. The parameter $\xi$ is the fraction of a $2\pi$ rotation applied in the rotations. For `rx', the same rotation angle is applied to all qubits. The `dim' results are the same but for simulations with limited entanglement, as explained in Section \ref{sec:ent}. The adaptive schedule of rotation angles is explained in section \ref{sec:adaptive}.}
\label{rxry}
\end{center}
\end{figure}

As mentioned above, the effect of an \texttt{x} or \texttt{y} gate on any qubit is to perform a reflection. Since \texttt{rx} and \texttt{ry} rotations correspond to partially performed \texttt{x} and \texttt{y} gates, these therefore correspond to a combination of the original image with the reflection. As such, it might be expected that these gates would result in the image becoming more symmetric. However, it is important to remember that the images are represented as quantum states during the application of the gates, and that different phases will be induced by the rotations on different terms within the superposition. The superposition of the states can therefore experience both constructive and destructive interference. The effect of this on the asymmetry is hard to predict, and so the naive expectation of reduced asymmetry cannot be expected to hold in general.

In order to see the effects of quantum superposition more clearly, we will construct an analogue of the Quantum Blur for which there is no interference. To see how this can be defined, note that a $\theta$ rotation of \texttt{rx} or \texttt{ry} results in a superposition of the original state and a bit flipped state. The magnitude of the amplitudes for these states are $\cos{\theta/2}$ and $\sin{\theta/2}$, respectively. If orthogonal, these would correspond to probabilities $\cos^2{\theta/2}$ and $\sin^2{\theta/2}$ of getting these states as outcomes if measuring in the appropriate basis. This means that, if we neglect any interference effects, the behaviour is equivalent to randomly applying a bit flip to the qubit with probability $\sin^2{\theta/2}$.

As such we can consider a decohered form of the superposition in which we imagine that a bit flip has randomly been applied or not, and take a corresponding weighted average with these probabilities of the flipped and unflipped images.

This form of the blur can certainly be expected to reduce the asymmetry. Indeed, we can expect to see a completely uniform image for $\theta = \pi/2$ where the flipped and unflipped images are evenly mixed.

Note that the implementation of the incoherent blur is done entirely classically. Only the weighting probabilities are calculated from the action of the quantum gates. However, it could also be implemented within a quantum simulation using bit flip noise with probability $\sin^2{\theta/2}$ on each qubit.

In Fig. \ref{rxry} we see that the incoherent effect has similar behaviour to the conventional blurs in the range of angles from $\theta=0$ (no effect) and $\theta=\pi/2$ (fully mixed). We find that the difference to the original image increases, and the asymmetry and detail decrease to zero. Over the entire range from $\theta=0$ to $\theta=\pi$ studied, periodic behaviour is seen as the weighting of the mixing moves from a bias against flipping images (with no flips at $\theta=0$) to a bias towards flipping (with the effect being the same as \texttt{x} or \texttt{y} gates on all qubits at $\theta=\pi$). For higher $\theta$ this would then go back to a bias against flipping (with no flips at $\theta=2\pi$), and so on.

The coherent effects of Quantum Blur show very different behaviour. Though the effects for an \texttt{rx} blur is identical to the incoherent case for $\theta=0$ and $\theta=\pi$, the interference effects prevent the vanishing asymmetry and detail at $\theta=\pi/2$. Instead, the asymmetry quickly jumps from the initial value of $0.15$ to around $0.1$ (decaying more quickly than the incoherent case) and mostly remains around this level while interference effects dominate.

For the effects of an \texttt{rx} blur on the image detail we see a small decay for small $\theta$, followed by mostly growth. The state detail, however, shows very similar behaviour to the asymmetry. The difference is due to the scrambling effect of the image at $\theta=\pi$: applying a flip to all qubits results in pixel values that where not previously neighbouring (in terms of coordinates) being moved together. The resulting discontinuities therefore increase the measured image detail. For the state detail this does not occur, since relationships between neighbouring strings remain invariant.

\begin{figure*}[htbp]
\begin{center}
\includegraphics[width=2\columnwidth]{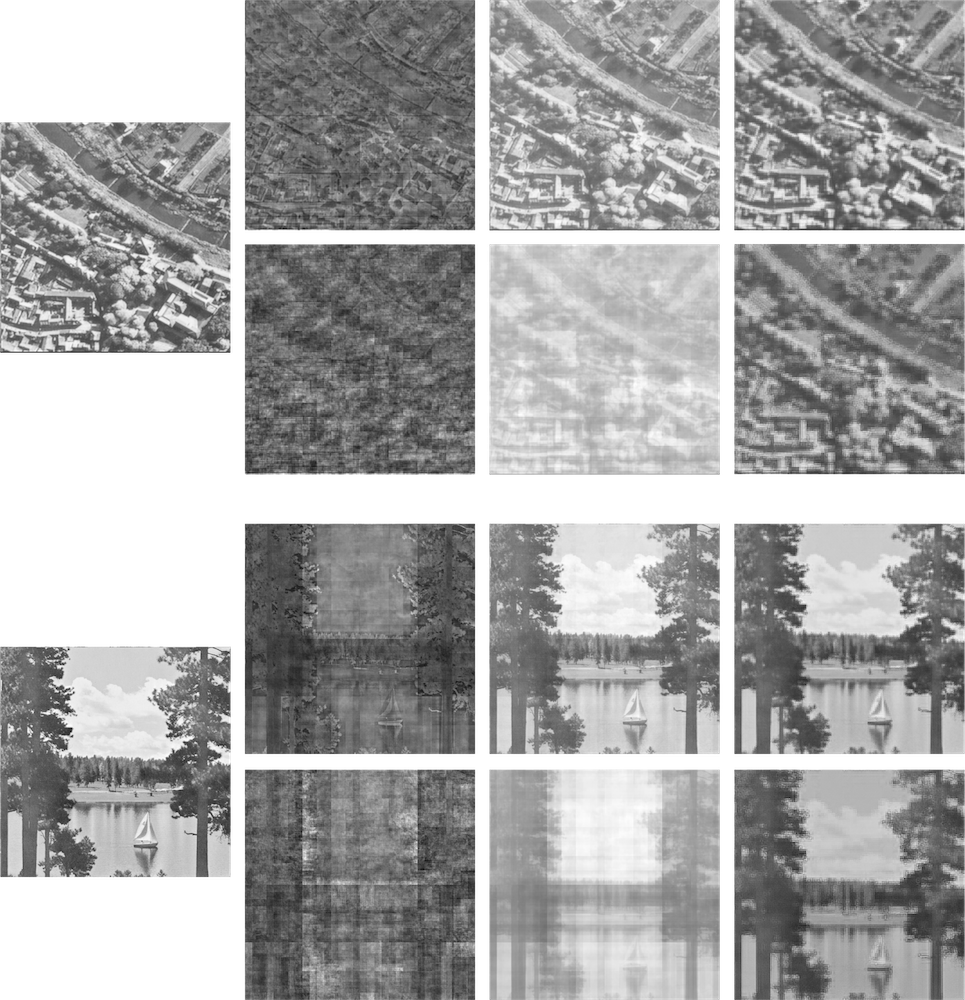}
\caption{Examples of some of the blur effects discussed, applied to two $512 \times 512$ pixel example images. From left to right, \texttt{rx} and incoherent blurs are shown for a given $\xi$ and the adaptive blurs is shown for $2\xi$. For each image the top line shows the effect for $\xi=16$ and the bottom for $\xi=8$.}
\label{examples}
\end{center}
\end{figure*}

Examples of the effects studied here applied to specific text images can be found in Fig. \ref{examples}.

\section{The Role of Entanglement} \label{sec:ent}

The preparation of images begins with the standard initial state of a quantum computer: $\left| 0 \right\rangle^{\otimes n}$. When measured, this outputs a bit string of all \texttt{0}s with certainty, and so corresponds to a single bright pixel at the corresponding coordinate. This is shown in Fig. \ref{dotdot} (a), where the bright pixel for the all \texttt{0} string is located at the top left. By applying gates we  `unfold' this single pixel into any desired image.

With single qubit rotations we can apply only reflections across the whole image. This is sufficient only to create simple shapes. For more complex images, it is necessary to apply operations only to specific regions. To address parts of an image we can use controlled operations, which would apply reflections only to those parts of the image whose coordinates correspond to the conditions on the control qubits. Since these operations are entangling, the circuits required to draw general images are necessarily entangling.

\subsection{Drawing a simple image}

As a simple example of this, we will consider the preparation of the a $32\times 32$ pixel version of the letter `I', with banding for additional detail. This requires 10 qubits, with 5 required to encode the x coordinate and 5 for the y coordinate. The initial state is $\left| 00000 \, 00000 \right\rangle$, where the space is used to separate the qubits for the x coordinate (on the left) from those for the y coordinate (on the right). For each, the qubits are numbered from 0-4, from right to left. We will refer to qubit 0 of the y register as $y-0$, and so on.

\begin{figure}[htbp]
\begin{center}
\includegraphics[width=0.95\columnwidth]{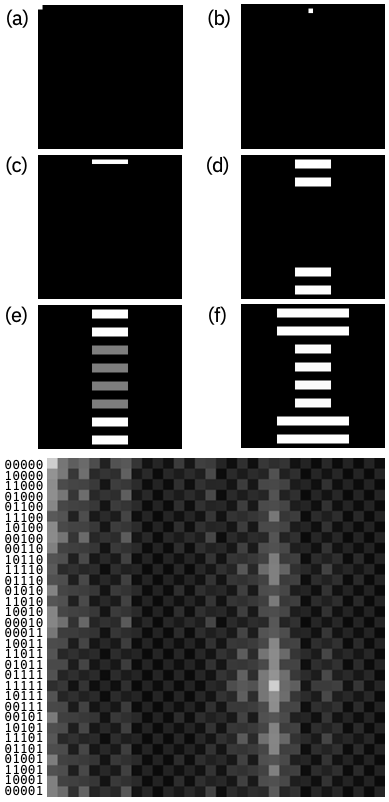}
\caption{The process of creating an `I' using quantum gates, as explained in the text. The bit string coordinates are shown for each y-axis value at the bottom. The x-axis coordinates are labelled in the same way from left to right. The example image here (of a Quantum Blur effect applied to an initial image of two pixels) is only to guide the eye.}
\label{dotdot}
\end{center}
\end{figure}

To begin, an \texttt{x} gate is applied on qubit $y-4$ to move the bright pixel to $(0,1)$, and then another on qubit $x-1$ moves it to $(15,1)$. The resulting image is shown in Fig. \ref{dotdot} (b).

Next, $\pi/2$ \texttt{rx} rotations are applied to qubits $x-0$, $x-3$ and $x-4$. These are reflections that expand the single point into the bar seen in Fig. \ref{dotdot} (c). A further $\pi/2$ \texttt{rx} rotation is then applied to qubit $y-3$ to widen the bar, and then the same to qubits $y-0$ and $y-2$ to copy it as shown in Fig. \ref{dotdot} (d). A further reflection applied to qubit $y-1$ then copies these bars into the center, as seen in Fig. \ref{dotdot} (e). This is done by a \texttt{rx} rotation with an angle of less than $\pi/2$, which results in the reflected bars having a smaller amplitude.

Finally, a reflection must be applied to extend the upper and lower bars. The can be done by a $\pi/2$ \texttt{rx} rotation on $x-2$. However, note that this would extend all the bars. To extend only the upper and lower, the rotation must be applied such that it only addresses the correct parts of the image. The parts in question are all such that qubit $y-1$ should be in state $\left|0\right\rangle$. The desired effect can therefore be applied by a \texttt{crx} gate, a controlled \texttt{rx}, conjugated by \texttt{x} gates on the control qubit. This applies the \texttt{rx} rotation to the target qubit only when the control is in state $\left|0\right\rangle$. Using $y-1$ as control and $x-2$ as target, this will extend only the upper and lower bars, as required. Note that the amplitude of these is then spread out over more points. For a suitably chosen angle for step (e), all points will be brought to equal brightness. This completes the image, as shown in Fig. \ref{dotdot} (f).

Since this image is mostly very simple, most of the operations applied here are on single qubits. Only the final operation requires a two qubit entangling gate. However, for more complex images, the need for entanglement will grow.

Also, recall that the standard method of encoding general images as states gives the same phase to all amplitudes. This would not have occurred in the state created above. In general, it may be possible to find ways to create images with less entanglement if there is not constraint on phases. Nevertheless, entanglement will be required in general.

\subsection{Effects of limiting entanglement}

\begin{figure*}[htbp]
\begin{center}
\includegraphics[width=1.8\columnwidth]{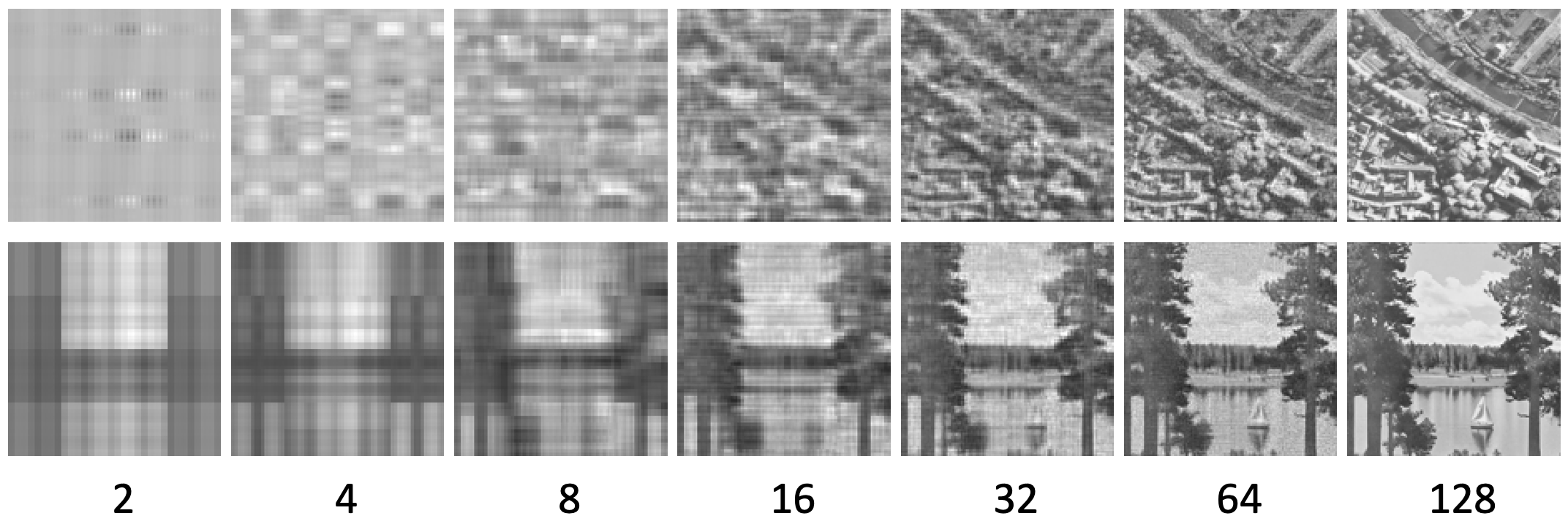}
\caption{Two $128 \times 128$ test images~\cite{images}, shown when reproduced on the Qiskit MPS simulator~\cite{qiskit} with different maximum bond dimensions.}
\label{entangle}
\end{center}
\end{figure*}

To demonstrate the necessity of entanglement, we can limit the amount of entanglement used when simulating the quantum process to see the effect on the final images. This was done by first transpiling the preparation operation to single and two qubit gates, and then running the resulting circuit on a matrix product state simulator. This simulator represents the state as a number of tensors connected by bonds whose dimension depends on how much entanglement exists between them~\cite{vidal-03,schollwoeck-11}. By placing an upper limit on the bond dimension, the amount of entanglement allowed to exist in the circuit can be limited. The effects of this are shown on two test images in Fig. \ref{entangle}.

For the sailboat image, certain aspects can be easily reproduced by single qubit reflections. Most prominently, these are the dark areas at either side representing the two trees and the shoreline. These aspects are therefore clearly visible even with very low entanglement. Features with fine detail, such as definition on branches and the sailboat, become visible as the amount of entanglement is increased.

For the image of a town, the mixture of different angles and high detail makes it difficult to create an approximation with single qubit gates. The image is therefore entirely unrecognizable until a large amount of entanglement is present.

The effects of limiting entanglement are also seen in Fig. \ref{rxry}, where the \texttt{rx} blur is shown with the bond dimension limited to 8 and 16. The effect on the difference to the original is to start off at a non-zero value (since the original cannot be represented), However, once the \texttt{rx} blur begins to take effect, the entanglement limited versions have a reduced effect. For the asymmetry and details, we find that the blurring effect of the entanglement limitation means that there is lower baseline level of these properties, and the effects induced by the \texttt{rx} blur are more muted. These results underline the fact that entanglement is required to see the full quantum-induced effects of the blur.

In particular, note that the asymmetry and details show a peak at $\theta=\pi/2$ for the \texttt{rx} blur. This peak comes at the point at which the interference effects are maximal. However, it is entirely absent from the two examples with limited entanglement, showing that the effects of interference cannot be fully represented in the images without entanglement.

Another way that the entanglement could be limited is through noise. A simple example of this would be dephasing noise applied after the image has been encoded. This would have the effect of destroying the entanglement, but would result in no observable difference if the image where decoded. In the most extreme form, this noise could remove the entanglement entirely, leaving only a classical probability distribution of bit strings.

Nevertheless, the noise would have large effects on the manipulation of the image. As a simple example of this, consider the encoding of a plain white image. This could be done using the unentangled state $\left|+\right\rangle^{\otimes n}$ or entangled states such as graph states, depending on the phase convention chosen. Which initial state is chosen greatly determines the way that any subsequent quantum gates affect the superposition state, and therefore the output image. However, complete dephasing would result in the maximally mixed state in all cases. This is invariant under any quantum gate, and so it would be impossible to alter the image in any way. The reduction of a quantum superposition to a classical probability distribution therefore reduces not only entanglement, but also the ability to implement effects due to quantum interference.

\section{Adaptive Blur Effect} \label{sec:adaptive}

The standard approach to an \texttt{rx} blur is to apply the same rotation angle to all qubits. However, for most images this will not create the most blur-like effect. For example, consider an image with just a single bright pixel. This pixel only has four neighbours, and so only rotations on four qubits will actually result in amplitude being exchanged with these neighbours. Any image larger than $4\times 4$ pixels will have more qubits than this. It would therefore be more effective to first determine the pertinent four qubits, and apply the rotation only to them.

In general, images are more complex than just a single bright pixel. Instead therefore will be many pixels with non-zero brightness, with four pertinent qubits for each. An overall weight for each qubit can then be determined based on how many pixels it affects, and the brightness of each. Then, rather than applying the same angle $\theta$ to every qubit, the full rotation could be applied only to the highest weighted qubit. For the rest, a proportion of the angle could be applied based on their weight. We refer to \texttt{rx} rotations applied adaptively applied in this way, as an the `adaptive blur'.

For a uniformly bright image, the lowest weight will go to qubits $x-0$ and $y-0$. These reflect around the middle, and so the only pixels it reflects onto their neighbours are those along the middle. For $x-1$ and $y-1$ there are two lines of reflection, and so twice as many pixels reflected to their neighbours. Similarly qubits $x-j$ and $y-j$ reflect $2^j$ as many pixels to their neighbours as $x-0$ and $y-0$. The weight assigned to each qubit will therefore double as we go up the registers.

This analysis will also be approximately true for any other sufficiently dense image. This exponential schedule of angles can therefore be used as an approximation of the adaptive blur, without the need for an analysis of the initial image.

In Fig. (\ref{rxry}) we see that the behaviour of the adaptive blur has similarities to both the coherent and incoherent cases. Like the former, the interference effects prevent the detail and asymmetry from vanishing. However, like the latter the decay is relatively slow and smooth. Also, the use of different angles for the \texttt{rx} rotations on different qubits means that the same periodic behaviour is not seen.

\section{Using Quantum Hardware}

Running the quantum circuits required for Quantum Blur can be done on real quantum hardware, or by emulation on conventional computing hardware. For most problems of interest for quantum computing, the former offers a significant advantage in computational complexity~\cite{nielsen-00}. However, since Quantum Blur was designed as a proof-of-principle method for relatively few qubits, this complexity advantage does not exist.

The emulation process can be performed with a complexity of at most $O(WH \log(WH)) = O(n 2^n)$ ~\cite{wootton-fdg}. Though this is exponential with the number of qubits, note that it is only a little higher than linear in the number of pixels. When the emulation method allows direct access to the statevector, the probabilities can be accessed directly from this with no additional complexity. When they must be sampled, the sampling complexity is the same as for real quantum hardware, as discussed below.

For running on real quantum hardware, there are three parts to consider: preparation of the state, running the blur circuit and sampling the output. For the first, a quantum circuit needs to be constructed that will prepare the initial state on the device. Since there are $WH$ values to be encoding, this circuit requires at least an $O(WH)=O(2^n)$ complexity in some form~\cite{iten-2016}. Typically this is a time complexity, though alternatives exist which place the burden on space complexity~\cite{araujo-21}. For the blur circuit, the complexity will depend on the complexity of the circuit. The \texttt{rx} blur, for example, can be implemented with the simultaneous application of\texttt{rx}, and so an $O(1)$ time complexity (and $O(n)$ space). Finally, for the sampling the process must be repeated enough times to sample the probabilities of each pixel with sufficient accuracy, which has a complexity of at least $O(WH) = O(2^n)$ for any sensible choice of accuracy. However, the coefficient in this case may be very high, depending on the accuracy required. The total complexity is therefore at least $O(WH)=O(2^n)$.

Note that since emulation avoids the step of constructing and running the initialization circuit, and can also avoid sampling of the output, the runtime can easily be very short in comparison with real quantum hardware. This is despite the slight complexity disadvantage for emulation. 

\begin{figure}[htbp]
\begin{center}
\includegraphics[width=0.85\columnwidth]{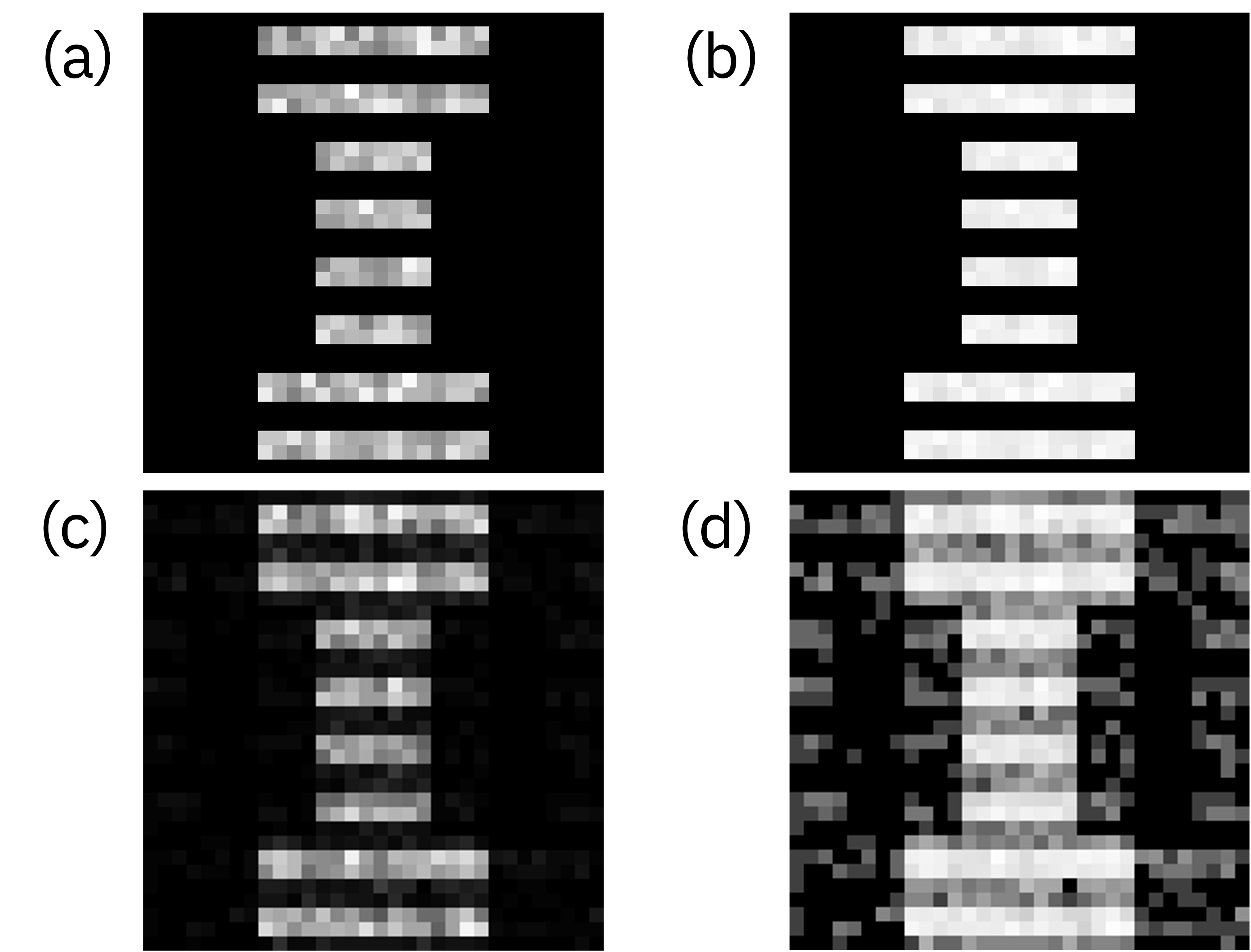}
\caption{A $32\times 32$ pixel `I' created through several different methods: (a) Run by emulation, but with the probabilities estimated from 8192 samples; (b) As before, but with the brightness scaling with the logarithm of the probabilities; (c) Sampled from 8192 runs on the 27 qubit \texttt{ibmq\_sydney} device; (d) as before, but logarithmically plotted.}
\label{real_run}
\end{center}
\end{figure}

The effects of of using real quantum hardware, and of sampling the probabilities, can be seen in Fig. \ref{real_run}. Note that the sampling causes statistical noise, which can be reduced by increasing the sample number. The use of current prototype quantum hardware also introduces the effects of imperfections in all quantum gates. This can be seen from the non-zero brightness in areas that should be dark. Since errors typically effect single qubits, they effectively induce additional reflections.

\section{Application to music}

Though Quantum Blur has mostly been used in the context of image manipulation, the technique is more general. This can be demonstrated with an application to music. To keep this in line with the previously discussed examples and to keep it as close to image manipulation as possible, we will first encode a musical score as an image, manipulating the image with Quantum Blur, and then converting it back to a musical score.

To convert a score to an image, we will use the y-axis to represent the semitone played and the x-axis to represent the beat at which it is played. For simplicity, we will consider only scores in which all notes have the same duration.

One complication in doing this is the fact that music commonly uses the 12-tone equal temperament system. Representing a score as an image could then be naturally achieved using 12 pixels in height for each octave. However, the encoding of images in Quantum Blur is most naturally achieved using powers of 2. When converting a score to an image we therefore transform first to a 16-tone system. When converting an image to a score we then similarly transform back to the standard 12-tone system.

\begin{figure*}[htbp]
\begin{center}
\includegraphics[width=1.5\columnwidth]{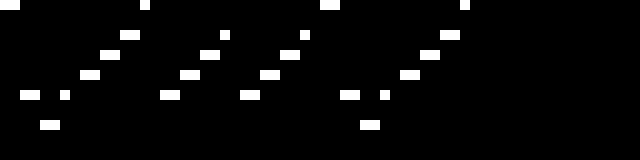}
\includegraphics[width=1.5\columnwidth]{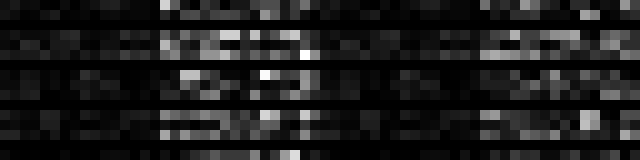}
\includegraphics[width=1.5\columnwidth]{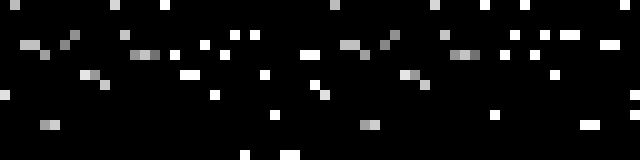}
\caption{Musical scores represented as images. The y-axis represents the tone of the notes on a 16-tone octave, with the lowest tones at the top. The x-axis represents the beat at which a note is played. The brightness of the pixels represents their volumes. The scores are `Twinkle, Twinkle, Little Star' with: (upper) no effect applied; (middle) \texttt{rx} on all qubits for $\theta = 0.41*\pi$; (lower) a cleaned version of the previous.}
\label{twinkle}
\end{center}
\end{figure*}

Such images are shown in Fig. \ref{twinkle}. These depict the image resulting from the score of `Twinkle, Twinkle, Little Star', the result of applying a strong blur effect to this original image, and a cleaned up version of the blurred image. The cleaned up image is used to remove the dissonant combinations of notes that would result if the score resulting from the blurred image would be played. This is done by retaining only the brightest pixel (or pixels) from each column. The brightness of these pixels in the cleaned up image is set to be the total brightness for their corresponding column in the blurred image, such that the total brightness of each column is also retained.

To hear the score resulting from this process, see \cite{soundcloud}. A Jupyter notebook detailing the process can be found at \cite{musichack}.

Note that this is intended only as a proof-of-principle use of Quantum Blur in music. Many variations of the encoding could be devised and used, leading to different results. For example, instead of fixed duration and variable volume one could use variable duration and fixed volume instead.

\section{Conclusions}

Blur effects typically have the effect of arbitrarily washing out a source image. However, as we have shown, the Quantum Blur effect retains a significant degree of asymmetry and detail. We have also shown that this behaviour emerges explicitly from the quantum phenomena that are crucial for quantum computation: superposition and entanglement. This underlines that the `Quantum Blur' is a genuine example of quantum software in use.

Nevertheless, it is important to note that such behaviour is not necessarily unique to the quantum approach. Qualitatively similar results could almost certainly be achieved through an entirely non-quantum approach. This work could therefore also serve as a motivation to seek out such methods.

Quantum Blur was created as an initial proof-of-principle example of quantum procedural generation. Though this work focuses on applications of Quantum Blur itself, it is worth noting that further methods for quantum procedural generation have been developed since. Particularly pertinent for music generation is the use of the same frameworks that underlie quantum natural language processing~\cite{karamlou-22,miranda-22}.

The source code for Quantum Blur can be found at \cite{quantumblur}. Examples of its use and other resources can be found at \cite{qisk-it}.

\section{Acknowledgements}
The authors would like to thank Roman Lipski for crucial discussions on his use of the method.
The authors would like to thank Elisa B\"aumer for comments on the manuscript. JRW acknowledges support from the NCCR SPIN, a National Centre of Competence in Research, funded by the Swiss National Science Foundation (grant number 51NF40-180604).

\bibliographystyle{unsrt}
\bibliography{references}

\begin{thebibliography}{10}

\bibitem{pcg}
Tanya~X. Short and Tarn Adams.
\newblock {\em Procedural Generation in Game Design}.
\newblock A. K. Peters, Ltd., USA, 1st edition, 2017.

\bibitem{lagae:10}
Ares Lagae, Sylvain Lefebvre, Robert~L Cook, Tony DeRose, George Drettakis,
  David~S Ebert, John~P Lewis, Ken Perlin, and Matthias Zwicker.
\newblock State of the art in procedural noise functions.
\newblock {\em Eurographics (State of the Art Reports)}, pages 1--19, 2010.

\bibitem{smith:11}
Adam~M. Smith and Michael Mateas.
\newblock Answer set programming for procedural content generation: A design
  space approach.
\newblock {\em IEEE Transactions on Computational Intelligence and AI in
  Games}, 3(3):187--200, 2011.

\bibitem{wootton-fdg}
James~R. Wootton.
\newblock Procedural generation using quantum computation.
\newblock In {\em International Conference on the Foundations of Digital
  Games}. {ACM}, September 2020.

\bibitem{capdeville-21}
F\'{e}lix Olart, Elo\"{\i}se Tassin, Laurine Capdeville, Luc Pinguet, Th\'{e}o
  Gautier, and Alain Lioret.
\newblock Quantum nodes: Quantum computing applied to 3d modeling.
\newblock In {\em ACM SIGGRAPH 2021 Posters}, SIGGRAPH '21, New York, NY, USA,
  2021. Association for Computing Machinery.

\bibitem{wootton-cog}
James~R. Wootton.
\newblock A quantum procedure for map generation.
\newblock In {\em 2020 IEEE Conference on Games (CoG)}, pages 73--80, 2020.

\bibitem{hamido-21}
Omar~Hamido Costa.
\newblock {\em Adventures in Quantumland}.
\newblock PhD thesis, UC Irvine, 2021.

\bibitem{schuld-20}
Maria Schuld, Alex Bocharov, Krysta~M. Svore, and Nathan Wiebe.
\newblock Circuit-centric quantum classifiers.
\newblock {\em Phys. Rev. A}, 101:032308, Mar 2020.

\bibitem{piispanen:23}
Laura Piispanen, Marcel Pfaffhauser, Annakaisa Kultima, and James~R. Wootton.
\newblock Defining quantum games, 2023.

\bibitem{clay}
Christopher Sciacca.
\newblock Upcoming video game will generate new levels using qiskit and a
  quantum simulator, 2020.

\bibitem{lipski:21}
Roman Lipski.
\newblock Making the invisible visible: A new exhibition of quantum art.
\newblock
  \url{https://medium.com/qiskit/making-the-invisible-visible-a-new-exhibition-of-quantum-art-ba1540b9ba82},
  2021.

\bibitem{lipski:23}
Roman Lipski.
\newblock This permanent other landscape.
\newblock
  \url{https://artsoftheworkingclass.org/exhibition/roman-lipski-this-permanent-other-landscape},
  2023.

\bibitem{libby:19}
Libby Heaney.
\newblock Art shared on twitter.
\newblock \url{https://twitter.com/LibbyHeaney/status/1187108052887113729},
  2019.

\bibitem{libby:20}
Libby Heaney.
\newblock Art shared on twitter.
\newblock \url{https://twitter.com/LibbyHeaney/status/1255043337138167808},
  2020.

\bibitem{lucal-59}
Harold~M. Lucal.
\newblock Arithmetic operations for digital computers using a modified
  reflected binary code.
\newblock {\em IRE Transactions on Electronic Computers}, EC-8(4):449--458,
  1959.

\bibitem{images}
USC-SIPI.
\newblock Image database - volume 3.
\newblock \url{https://sipi.usc.edu/database/database.php?volume=misc}.
\newblock Accessed: 2021-11--9.

\bibitem{opencv_library}
G.~Bradski.
\newblock {The OpenCV Library}.
\newblock {\em Dr. Dobb's Journal of Software Tools}, 2000.

\bibitem{vidal-03}
Guifr\'e Vidal.
\newblock Efficient classical simulation of slightly entangled quantum
  computations.
\newblock {\em Phys. Rev. Lett.}, 91:147902, Oct 2003.

\bibitem{schollwoeck-11}
Ulrich Schollwöck.
\newblock The density-matrix renormalization group in the age of matrix product
  states.
\newblock {\em Annals of Physics}, 326(1):96--192, 2011.
\newblock January 2011 Special Issue.

\bibitem{qiskit}
The Qiskit~Community.
\newblock Qiskit: An open-source framework for quantum computing, 2021.

\bibitem{nielsen-00}
Michael~A. Nielsen and Isaac~L. Chuang.
\newblock {\em Quantum Computation and Quantum Information}.
\newblock Cambridge University Press, 2000.

\bibitem{iten-2016}
Raban Iten, Roger Colbeck, Ivan Kukuljan, Jonathan Home, and Matthias
  Christandl.
\newblock Quantum circuits for isometries.
\newblock {\em Phys. Rev. A}, 93:032318, Mar 2016.

\bibitem{araujo-21}
Israel~F. Araujo, Daniel~K. Park, Francesco Petruccione, and Adenilton~J.
  da~Silva.
\newblock A divide-and-conquer algorithm for quantum state preparation.
\newblock {\em Scientific Reports}, 11(1):6329, 2021.

\bibitem{soundcloud}
James Wootton.
\newblock Quantum twinkle 2021.
\newblock
  \url{https://soundcloud.com/james-wootton-348392631/quantum-twinkle-2021},
  2021.

\bibitem{musichack}
James Wootton.
\newblock Quantum music hack.
\newblock
  \url{https://github.com/quantumjim/QuantumMusicHack/blob/main/README.md},
  2021.

\bibitem{karamlou-22}
Amin Karamlou, James Wootton, and Marcel Pfaffhauser.
\newblock Quantum natural language generation on near-term devices.
\newblock In {\em Proceedings of the 15th International Conference on Natural
  Language Generation}, pages 267--277, Waterville, Maine, USA and virtual
  meeting, July 2022. Association for Computational Linguistics.

\bibitem{miranda-22}
Eduardo~Reck Miranda, Richie Yeung, Anna Pearson, Konstantinos Meichanetzidis,
  and Bob Coecke.
\newblock {\em A Quantum Natural Language Processing Approach to Musical
  Intelligence}, pages 313--356.
\newblock Springer International Publishing, Cham, 2022.

\bibitem{quantumblur}
James Wootton and Marcel Pfaffhauser.
\newblock Quantum blur.
\newblock \url{https://github.com/qiskit-community/QuantumBlur}, 2021.

\bibitem{qisk-it}
James Wootton.
\newblock Links to resources about quantum blur.
\newblock \url{https://qisk.it/quantumblur}, 2023.

\end{thebibliography}

\end{document}